\documentclass{article}
\usepackage[preprint]{icml2026}

\usepackage{microtype}
\usepackage{graphicx}
\usepackage{amsmath,amssymb}
\usepackage{booktabs}
\makeatletter
\let\algorithmic\@undefined
\let\endalgorithmic\@undefined
\let\algorithmicindent\@undefined
\makeatother
\usepackage{algpseudocode}
\setcitestyle{numbers,square,comma}
\icmltitlerunning{Enhancing the MADDPG Algorithm for Multi-Agent Learning}

\begin{document}
\onecolumn

\icmltitle{Enhancing the MADDPG Algorithm for Multi-Agent Learning via Action Inference and Importance Sampling}

\begin{icmlauthorlist}
\icmlauthor{Marc Walden}{ucla}
\icmlauthor{Jason Liu}{ucla}
\icmlauthor{Ryan Liu}{ucla}
\icmlauthor{Hamza Khan}{ucla}
\icmlauthor{Shaashwath Sivakumar}{ucla}
\end{icmlauthorlist}

\icmlaffiliation{ucla}{Department of Mathematics, University of California Los Angeles, Los Angeles, CA, USA}

\icmlcorrespondingauthor{Marc Walden}{marcwalden@g.harvard.com}

\printAffiliationsAndNotice{}
\vspace{1em}

\begin{abstract}
 We investigate multi‐agent deep reinforcement learning and propose two enhancements to the Multi‐Agent Deep Deterministic Policy Gradient (MADDPG) algorithm. First, we introduce a novel Action Inference mechanism that enables each agent to predict other agents' intended actions, thereby improving the accuracy and stability of its own policy. Second, we apply an importance sampling strategy, using geometric distribution, in the replay buffer to prioritize more recent and informative experiences, which helps mitigate the non‐stationarity inherent in multi‐agent environments. We evaluate both modifications on the discrete‐action Predator–Prey task provided by the PettingZoo library~\cite{farama2020}, a flexible Python interface for general multi‐agent reinforcement learning benchmarks. Our results indicate that Action Inference is effective in improving learning stability and inter-agent cooperation and that importance sampling using geometric distribution can lead to significant improvements in exploration efficiency over standard MADDPG. Code available at \url{https://github.com/shaashwathsivakumar/MARL_Proj} 
 \end{abstract}

\section{Introduction}
Multi-agent reinforcement learning (MARL) is a rapidly growing area in artificial intelligence, involving multiple agents that interact with each other in cooperative and competitive scenarios within a shared environment. MARL has become increasingly relevant to applications such as robotics, autonomous vehicles, and strategic game playing, which require agents to learn effective policies from experience. In this project, we examine applications where each agent operates under partial observability: at every time step, it receives only a subset of the full environment state.

Several approaches have been proposed to tackle reinforcement learning in multi-agent domains. Tan (1993) applied traditional one-step Q-learning to MARL problems \cite{tan1993multi}. However, independent Q-learning typically fails in these settings due to environmental non-stationarity: from any individual agent’s perspective, the policies of its peers evolve over time in ways that its own updates cannot capture. Policy-gradient methods such as REINFORCE also struggle in MARL, since their gradient estimators exhibit variance that grows with the number of agents—resulting in high computational cost and poor convergence.

Actor-critic algorithms mitigate the high-variance issue by introducing a learned value-function baseline, which reduces variance without biasing gradient estimates. The critic network provides a more stable estimate of expected returns, allowing the actor to apply more reliable updates and thereby improving learning stability.
 
Multi-Agent Deep Deterministic Policy Gradient (MADDPG) \cite{lowe2020} extends deterministic policy gradients to multi-agent settings by combining centralized training with decentralized execution. During training, each agent’s critic has access to the full joint state and all agents’ actions; during execution, each actor selects actions based solely on its local observation.

To enhance learning stability and coordination, we introduce two extensions to the standard MADDPG framework. First, we implement a novel action inference mechanism that enables each agent to predict its peers’ previous actions and incorporate these predictions into the critic’s inputs during training. Second, we replace uniform replay sampling with a geometric sampling strategy, which prioritizes more recent experiences in the buffer to further mitigate non-stationarity.

We evaluate the baseline MADDPG algorithm, the action inference augmentation, and the geometric sampling method using the Predator–Prey game from the Simple Tag suite in the PettingZoo library \cite{farama2020}. In this environment, a team of predators cooperates to capture a prey agent under partial observability; discrete action choices and environmental hazards such as boundaries produce a mixed cooperative–competitive dynamic.

For the initial implementation of standard MADDPG, we adapted code from an open-source repository by Git-123-Hub \cite{git123hub2021}. We then extended this codebase to integrate our action inference and geometric sampling methodologies.

\section{Background}

\subsection{Reinforcement Learning Framework}
In reinforcement learning (RL), agents learn to make sequential decisions by interacting with an environment \(\mathcal{E}\) over discrete time steps \(t=0,1,2,\dots\).  At each step, agent \(i\) observes \(o_t^i \in \mathcal{O}\), selects an action \(a_t^i \in \mathcal{A}\), and receives a scalar reward \(r_{t+1}^i \in \mathbb{R}\), as seen below in Figure 1. The agent’s goal is to discover a policy \(\pi_{\theta_i}(a^i \!\mid o^i)\), parameterized by \(\theta_i\), that maximizes the expected discounted return
\begin{equation}
J(\theta_i) \;=\; \mathbb{E}\Bigl[\sum_{t=0}^\infty \gamma^t\,r_{t+1}^i\Bigr],
\end{equation}
where \(\gamma\in[0,1)\) is the discount factor that reduces the value of future rewards.  Observation spaces \(\mathcal{O}\) may be high-dimensional (e.g., images, LIDAR scans) or low-dimensional (e.g., positions and velocities), and action spaces \(\mathcal{A}\) can be discrete (a finite set of moves) or continuous (real‐valued control signals). By interacting with other agents and the environment during training, the goal is for each agent to learn a policy that maximizes the value of Equation (1). The most common approach in reinforcement learning for optimizing policies is General Policy Iteration (GPI), which alternates between policy evaluation and policy improvement until convergence. Policy evaluation estimates the action-value function, mapping each state to the expected return under the current policy, and updates it to better reflect the policy’s performance. Policy improvement uses the updated value function to refine the policy, typically by selecting actions that yield higher expected returns, thereby making the policy more greedy with respect to the current value estimates.

\begin{figure}[h]
    \centering
    \includegraphics[width=0.5\linewidth]{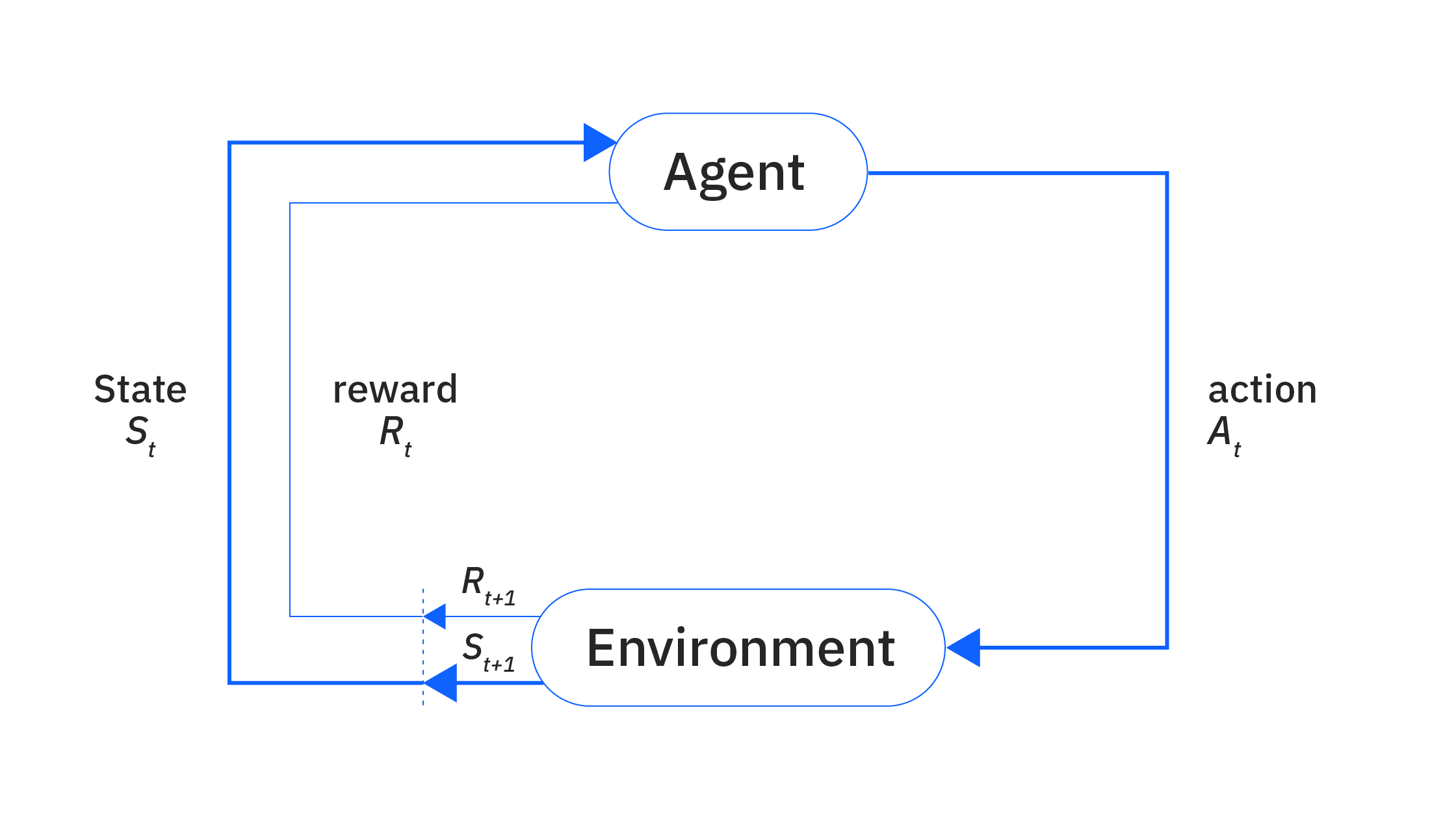}
    \caption{The Reinforcement Learning framework.}
    \label{fig:rl-pic1}
\end{figure}

\subsection{Deep Q-Learning and Multi-Agent Extensions}
A seminal advance in deep RL was the Deep Q-Network (DQN) \cite{mnih2013}, which approximates the action‐value function \(Q(s,a)\) with a deep neural network. This value can be recursively written, and thus DQN learns by minimizing the Bellman error:
\begin{equation}
\mathcal{L}(\phi) \;=\; \mathbb{E}_{(s,a,r,s')\sim\mathcal{D}}\Bigl[\bigl(r + \gamma\,\max_{a'} Q_{\phi^-}(s',a') - Q_{\phi}(s,a)\bigr)^2\Bigr].
\end{equation}
The key feature of DQN is that it uses an experience replay buffer to store the agent's interactions with the environment that are later sampled for training. While DQN excels in single‐agent tasks, Q‐learning becomes unstable in multi‐agent settings due to non‐stationarity: each agent’s policy updates are effectively changing the environment dynamics for others.

\subsection{Policy-Gradient Methods and MADDPG}
Policy-gradient methods directly parameterize the policy \(\pi_\theta(a \mid o)\) and optimize it by maximizing a performance objective \(J(\theta)\). The policy is updated via gradient ascent:

\[
\theta \leftarrow \theta + \alpha \nabla_\theta J(\theta)
\]

where \(J(\theta) = \mathbb{E}_{\pi_\theta}\big[\sum_t \gamma^t r_t\big]\) is the expected return under the policy. By the Policy Gradient theorem ~\cite{sutton2018}, we can obtain the gradient for the update: 

\begin{equation}
\nabla_\theta J(\theta) \;=\; \mathbb{E}\!\bigl[\nabla_\theta \log\pi_\theta(a\!\mid o)\,Q^{\pi}(o,a)\bigr].
\end{equation}
The Multi-Agent Deep Deterministic Policy Gradient (MADDPG) algorithm \cite{lowe2020} extends deterministic policy gradients to multi-agent domains by giving each agent \(i\) an actor \(\mu_{\theta_i}(o_i)\) and a centralized critic \(Q_{\phi_i}(x,a_{1:N})\) for training, where \(x\) is the joint state (or observations) and \(a_{1:N}\) the joint actions.  Thus, applying centralized training to environments requiring decentralized execution mitigates the non-stationarity problem by allowing critics full access to joint observations and actions during learning. Through this, the critic guides the actor to implicitly  learn features from the agent's decentralized observation that are informative of the centralized state, enabling better action selection. Both the actor and critic are trained alternately using samples from a replay buffer. The critic learns an approximation to the on-policy joint action–value function induced by the current policies, based on empirical interaction with the environment. The actor then updates its policy via deterministic policy gradients computed through the learned critic, performing gradient ascent on the expected return from the initial state distribution. The classic MADDPG training loop is summarized in Algorithm~\ref{alg:maddpg} (see appendix). 
\\

\section{Methodology}

\subsection{Standard MADDPG Implementation}
First, we translated classic MADDPG (Algorithm~\ref{alg:maddpg}) into PyTorch and trained for \(3 \times 10^4\) episodes of 25 steps. We opted for a discrete implementation of the action space since we are only working in 2 dimensions. The possible discrete actions include up, down, left, right, and no movement. We used the default design of the state and observation space of each agent from the "simple\_tag\_v3" environment from the PettingZoo library, with parallel execution and all settings as default. We implemented the state using a tensor of shape \((B,X)\) and joint actions \((B,N,A)\), which we flatten the joint actions and join them with the state to make \((B,N \times A + X))\) as input for the centralized critic, where B represents the batch-size, N represents the number of agents, X represents the size of the global state, and A represents the size of the joint actions. The critic networks use two hidden layers of 64 ReLU units, and the actor networks use three layers of 64 ReLU units. All networks are trained using the Adam optimizer with a learning rate of 0.01, a batch-size of 1024, and the Xavier initialization. The parameters of the target network are updated gradually using a soft update to stabilize learning at a rate of  \(\tau=0.02\). In order to ensure sufficient variance over the initial replay buffer, prior to learning, we generate the first 2000 episodes using uniformly random sampled actions.

\subsection{Action Inference}
Action Inference improves MADDPG by enabling each agent to predict both its collaborators’ and competitor's last actions using a neural network. $\textbf{AI\_Net}()$ is a neural network that takes in: $\mathbf{o_i}$, the current observation, and $\mathbf{o_i^-}$, the previous observation, for agent $i$; and outputs: $\mathbf{\hat{a}_i}\in \mathbb{R}^{N \times|\text{a\_space}|}$, an estimate of the joint action vector: $\text{concat}(\text{onehot}(a_1),\dots,\text{onehot}(a_N))$. 
\\

A straightforward implementation—training a separate network for each agent to predict the past actions of all agents—would incur significant computational cost, requiring a total output size of: $N \times (N \times |A|)$. This would put the process at $O(N^2)$ before even considering the significant overhead that comes with training and executing the much deeper and larger networks that are required as $|A|$ increases. Therefore, even if doing so was to prove useful within low-dimension testing environments, it would likely prove unable to be effectively scaled up to larger applications.
\\

However, in applications where the observation structure allows the parameters to be cleanly partitioned into $N$ subsets (one per agent), we can leverage a novel architecture to cleverly mitigate the aforementioned computational cost. Instead of using a large network for each agent, we can take advantage of the inherent separability of the observation by only training a few generalized modules corresponding only to each unique observer-observed pair of agent types. Then, each module is only responsible for taking in a small "bundle" of data (routed from the appropriate subset of the observer's observation)  pertinent to an individual agent $k$ (of the observed agent type) and generates $\mathbf{\hat{a}_{i,k}}$ to predict $\text{onehot}(a_k)$. We will refer to these individual modules as Directional Social Awareness Module(s) for those pertaining to observing other agents, and Directional Self-Awareness Module(s) for those pertaining to inferring the agent's own past actions (more on the purpose of that later). On runtime, the overall AI\_Net manager, according to a subset key, forms and delivers the necessary bundles to parallel instances of each module, returning $\mathbf{\hat{a}_{i,k}}$, which is then concatenated in order to construct $\mathbf{\hat{a}_i}$, agent $i$'s estimate of the joint action vector. 

\begin{figure}[H]
    \centering
    \includegraphics[width=0.9\linewidth]{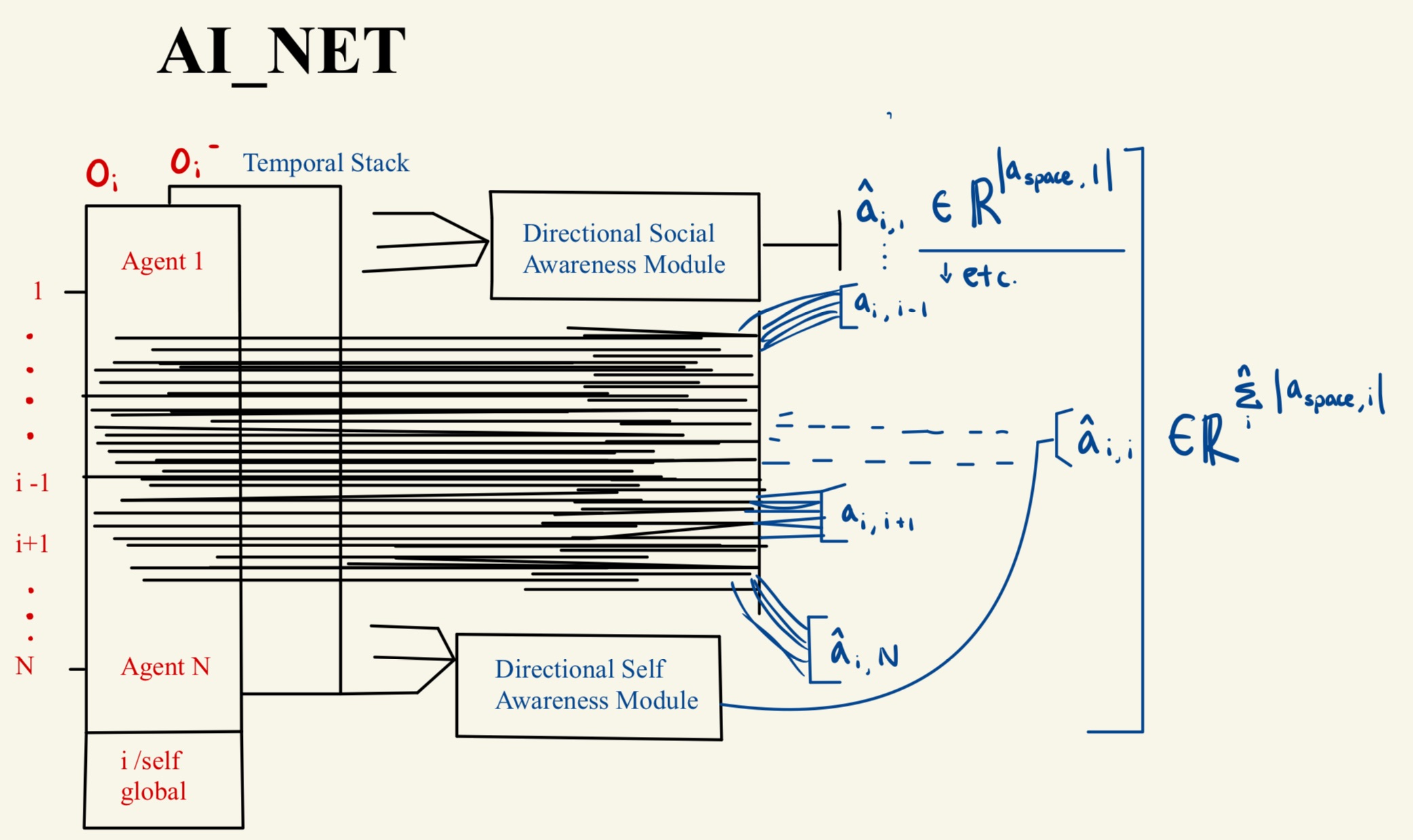}
    \caption{\textbf{Diagram of AI\_Net Architecture for Separable Observations}: Each pair of temporally stacked observation parameters undergoes an arithmetic difference operation, creating a third, equal-length set of related parameters. These differences alongside the sets of original parameters are grouped by agent and fed into a corresponding instance of the Directional Social Awareness Module, joined by parameters related to the observing agent, $i$, and anything indicative of reference frame or global behavior, to each compute a subset of the overall $\mathbf{\hat{a}_i}$. Parameters relating to the observing agent, $i$, and anything indicative of reference frame or global behavior are fed into the Directional Self Awareness Module to compute and insert the subset representing the agent's proprioception (which, if capable of being distorted, can engage recurrent sensorimotor loops that encode and refine motor memory).}
    \label{fig:example}
\end{figure}

Note that one of the many advantages that comes with this approach is that since the input consists of temporally stacked observations, we can compute the temporal difference for each stacked parameter pair. This itself functions as an important parameter, which the much smaller modules are now able to accommodate.
\\

We pre-train \textbf{AI\_Net}() on a diverse, representative sample (typically simulated from randomly generated or a large library of played episodes) to enable it to accurately predict the actions that led to transitions between observation pairs for any situation the agents may experience. For reference in the full pseudocode implementation of $\textbf{AI\_Net}()$ is available as Algorithm \ref{alg:PTAI Training w/ Sep.ble Obs} in the appendix. 
\\

Thus, within the new MADDPG training loop, the modified actor, $\mu_i$, takes in $\mathbf{\hat{a}_i}$ as a parameter (in addition to the standard $o_i^t$), and this psuedocode is shown as Algorithm \ref{alg:maddpg w/ PTAI} in the appendix. Since AI\_Net is already fully trained, the process that calculates it is completely stationary, not contributing to any additional complexity within the loss function or optimization process. This method, for almost no computational cost, improves on MADDPG by providing the agents with input parameters that inherently represent factors explaining what would otherwise be perceived as "non-stationarity" within the environment and enabling a mode of implicit communication, allowing them to learn to collectively influence the environment together.

We implemented both the AI\_Net manager and the individual Awareness modules as classes within Python using the PyTorch library. Everything is handled as scalable tensors capable of computing over batches, offloading most work and storage to the accelerator, leaving the CPU only to schedule.

\subsection{Importance Sampling Method}
In the standard MADDPG algorithm, transitions—each comprising a state, action, reward, and next state—are stored in a replay buffer and sampled uniformly for training. This uniform sampling can under-represent critical experiences, particularly in non-stationary multi-agent environments. To mitigate this, we bias sampling toward more recent interactions by drawing indices from a geometric distribution. The geometric distribution models the number of Bernoulli trials required to achieve the first success. When used for replay-buffer indexing, it assigns higher probability to more recent entries. Its probability mass function is as follows:

\[
P(X = k) = (1 - p)^{k - 1} p, \quad \text{for } k = 1, 2, 3, \ldots
\]

where \(p\) represents the per-trial success probability. Higher values of \(p\) produce thinner tails, and thus a stronger recency bias. Figure 3 illustrates the resulting sampling distribution for a replay buffer of size \(7.5 \times 10^5\), showing a pronounced preference for the most recent transitions (where the geometric distribution values are used as reverse indices for the replay buffer).

\begin{figure}[h]
    \centering
    \includegraphics[width=0.7\linewidth]{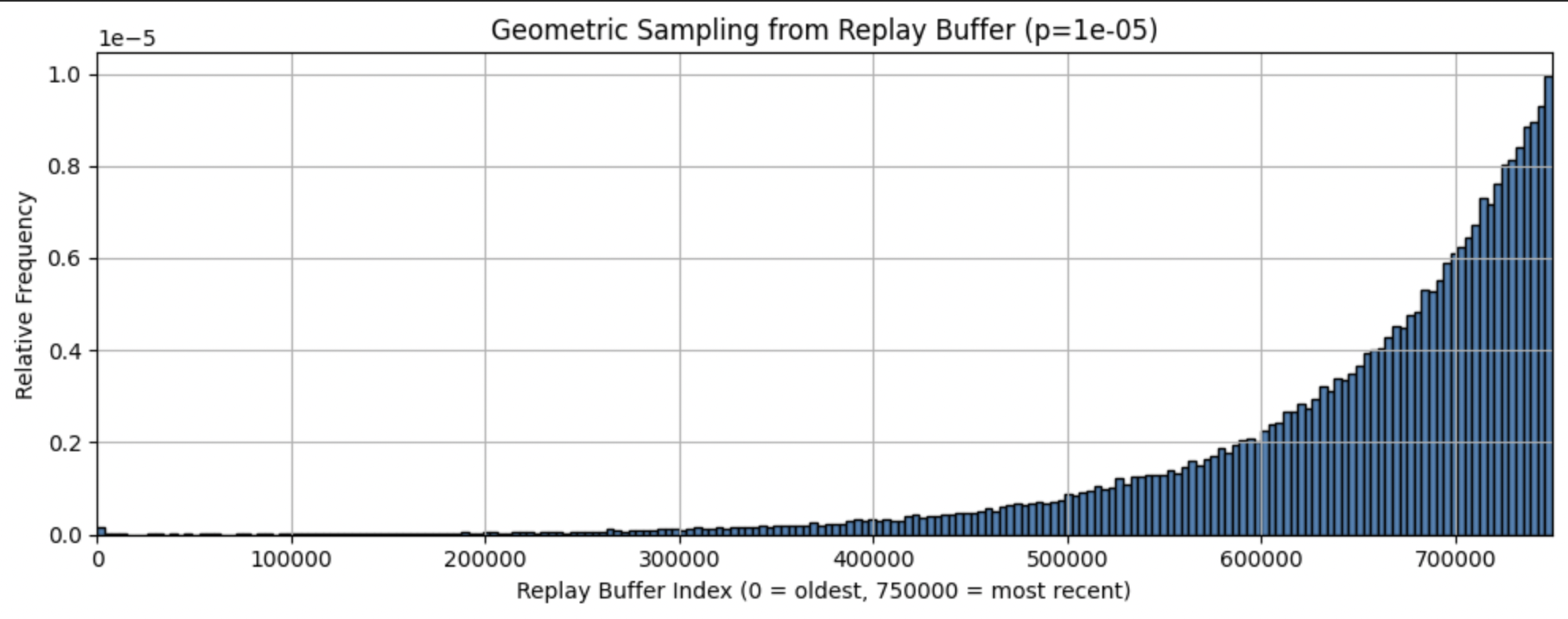}
    {\caption{Graph of the relative frequency of sampling from a replay buffer of size \(7.5 \times 10^5\) using a geometric distribution with parameter \(p = 10^{-5}\).}}
    \label{fig:geom}
\end{figure}

\subsection{Experimental Setup}
To quantify the results of our theoretical gains, we design an experiment to measure relative differences in exploration efficiency between the methods.
\\

First, we train both the predator and the prey using the standard MADDPG model. This creates an advanced prey actor, capable of dodging and weaving between the predators at high speed. For future runs, we then import the existing actor network for our prey to be used as a constant within our experiment, and only train the predators. This effectively makes the pre-trained prey part of the environment, resulting in an entirely cooperative environment for the agents. The prey, having already been significantly trained against capable adversaries, poses itself as a very difficult target to tag, so any benefits inherent to our methods would arise in how quickly and effectively the predator agents are able to initially coordinate to devise strategies leading to reward.
\\

To compare these methods, we initialize three distinct monitored training runs for the predator agents: one with standard MADDPG, one with Pre-Trained Action Inference (PTAI), and one with geometric importance sampling (Geom). Therefore, by plotting the monitored training data of the different methods against one another, we can compare any differences in early-run exploration efficiency.

\section{Results}
 
\subsection{Evaluation Metrics}
We evaluate each model based on the following metrics:
\begin{itemize}
    \item \textbf{Cumulative Max per Episode:} The max reward achieved by agent in one episode
    \item \textbf{Moving Average per Episode:} The average reward achieved by agents over the last 600 episodes 
\end{itemize}

\begin{figure}[h]
    \centering
    \includegraphics[width=1.0\linewidth]{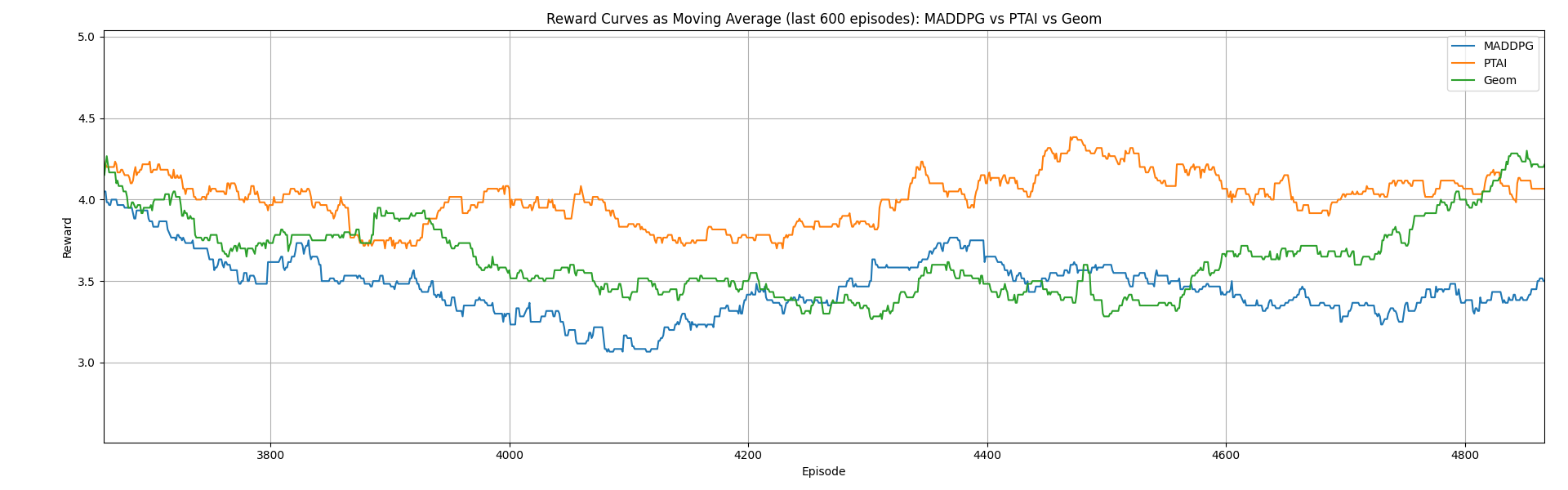}
    {\caption{Graph of the Moving Average (last 600) by Episode for each Method.}}
    \label{fig:mov_avg}
\end{figure}

\begin{figure}[h]
    \centering
    \includegraphics[width=0.9\linewidth]{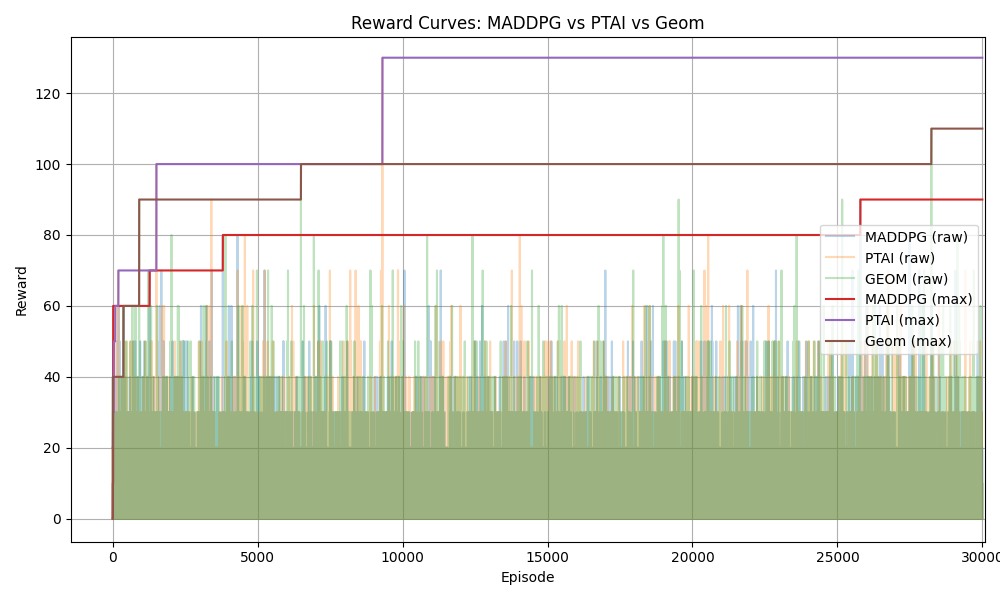}
    {\caption{Graph of the Cumulative Max and Raw Rewards by Episode for each Method.}}
    \label{fig:cum_max}
\end{figure}

In Figure 4, all methods exhibit indistinguishable behavior for the first 2,000 episodes, since during that period agents explore exclusively by sampling actions uniformly at random. This exploration strategy causes the replay buffer to be initially dominated by the resulting temporal-difference steps. Transitions generated by the learned policies only gradually begin to occupy a significant portion of the buffer, and even for the fastest variant this shift does not become pronounced until approximately 3,600 episodes
\\

 Cumulative maximum reward shown in Figure 5 is a useful metric for comparing exploration efficiency because consistently higher values indicate faster discovery of high-reward strategies. If all methods were equally efficient—meaning episode rewards were drawn from the same underlying distribution—then the identity of the algorithm achieving the current maximum would fluctuate randomly over time. However, Figure 3 shows a stable ordering of maximum rewards: the pre-trained action inference variant (PTAI) attains the highest values, the geometric-sampling method ranks second, and the standard MADDPG remains lowest. This persistent ranking suggests that the methods differ significantly in how quickly they uncover high-reward behaviors.
\\

As indicated in Figure 4, monitoring the tradeoffs in Reward by Episode, the PTAI method sets itself apart quickly from the beginning by consistently producing overall higher reward values, steadily dominating the other methods. Throughout this, Geom gradually joins the mix, beginning to consistently match values within the range of PTAI, and upon later runs, Geom actually appears to possibly dominate PTAI. We rarely see standard MADDPG's values individually exceed those of the other variations.
\\

This indicates that PTAI trains towards consistently good rewards, even from quite early. Geom, upon building a large enough replay buffer, will occasionally see its policies get pushed in the desired direction of larger cumulative reward, leading it to dominate for a time. Classic MADDPG appears slower than both, with its typical behavior barely exceeding even when Geom's samples fail to push it in the right direction.
\\

The stability of each method is shown within Figures 4. Both Geom and standard MADDPG have behavior that oscillates due to their random sampling combined with any non-stationarity inherent to their view of the environment. While these waves initially center around a similar value, the peaks of Geom can occasionally lift off, bringing up the overall wave shape with it. Thus, it appears that importance sampling from a geometric distribution provides a beneficial improvement over standard MADDPG since it allows it to immediately capitalize on good runs that reveal critical information informing better policies, with little to no sacrifice to its overall stability. PTAI learns faster and more steadily than the other methods, likely because Action Inference provides the agents with more useful information on the environment. This leads to smoother policy updates, avoiding the back-and-forth progress that MADDPG and Geom tend to exhibit.
\\

\section{Discussion} 
\subsection{MADDPG with Action Inference}
MADDPG augmented with pre-trained action inference (PTAI) achieved the highest performance in our experiments. The action inference, implemented as a pre-trained network that predicts each agent’s previous action, supplies an additional input feature to the actor network. Incorporating this feature accelerates policy convergence by providing richer context about state transitions. Intuitively, knowledge of the immediate past action will disambiguate observations that would otherwise appear similar, leading to more informed decision-making. Importantly, this enhancement did not introduce overfitting; because the previous action is directly relevant to predicting future outcomes, it generalizes well across episodes. Given its clear performance benefits, we recommend integrating PTAI into various other multi-agent reinforcement learning applications.
\\

\subsection{MADDPG with geometric distribution sampling}
The geometric-sampling variant of MAADPG also performs better than the baseline MADDPG implementation, highlighting a promising approach for biasing replay towards recent experiences. However, we believe there may be more room to further optimize the geometric-decay parameter \(p\) which we were unable to explore due to computational and scheduling constraints. In future work, we plan to investigate strategies for dynamically tuning \(p\) during training to test for better performance when sampling. We also believe it could be insightful to experiment with different sampling distributions (e.g., exponential) since a more balanced selection of experiences could also play a factor in performance. Overall, our implementation of geometric sampling serves as a valuable proof-of-concept, opening multiple avenues for enhancing experience prioritization in multi-agent reinforcement learning.

\subsection{Standard MADDPG}
We utilized a standard MADDPG implementation as a baseline model to evaluate the performances of our two proposed methodologies. It converged more slowly than the action inference variant, since it lacked the access to peers' predicted actions, thus causing it to require more interactions to learn coordinated behaviors. Towards the beginning of its training, our baseline model performed comparably to the geometric sampling implementation. However, at a larger episode count, the geometric sampling model surpasses the baseline MADDPG model as well, allowing it to converge quicker, as shown in Figure 4.

\section{Conclusion}

We extended the standard MADDPG algorithm with two key enhancements: incorporating a pre-trained action inference method allowed agents to predict and leverage their peers’ previous actions, and an alternative sampling mechanism from the replay buffer allowed us to train using more recent data points, thus mitigating the problem of non-stationarity. Evaluated on PettingZoo’s predator-prey environment from the perspective of the predator agents, our action inference variant significantly accelerated convergence and improved cooperative behavior. Our geometric sampling method also improved convergence and served as a proof-of-concept for the potential benefits of prioritizing recent experiences in multi-agent training. Together, these results highlight the effectiveness of combining action inference and recency-biased sampling to improve learning stability and coordination in multi-agent reinforcement learning.

\section{Acknowledgments}

We would like to express our sincere gratitude to Professor Tao Gao for his invaluable guidance throughout this project.

Our work began with a study of the foundational principles of reinforcement learning, as presented in the textbook by Sutton and Barto~\cite{sutton2018}. Following Professor Gao’s recommendation, we explored various implementations of deep reinforcement learning, starting with single-agent methods such as those discussed by Kurin et al.~\cite{kurin2021}. This exploration led us to the domain of multi-agent reinforcement learning, drawing from the works of Yu et al.~\cite{yu2021ppo} and Chen et al.~\cite{chen2016decentralized}.

We were particularly inspired by the project \emph{A Flock of Rogue Drones}~\cite{roguedrones2020}, which motivated our decision to apply multi-agent reinforcement learning to a tag-based environment. In searching for related implementations, we discovered the environment developed by Lowe et al.~\cite{lowe2020}, later released by OpenAI~\cite{openai2017}.

As the original environment depended on deprecated libraries, we transitioned to the updated PettingZoo implementation maintained by the Farama Foundation~\cite{farama2020}. Since this version used a random policy, we integrated the MADDPG algorithm by adapting an existing codebase from Git-123-Hub~\cite{git123hub2021}, modifying it to suit the specific needs of our project.

\bibliography{references.bib}
\bibliographystyle{plainnat}

\section*{Appendix}

We include pseudocode for the three relevant algorithms used in this paper below. These namely include the standard MADDPG algorithm, our pre-trained action inference, and the incorporation of our action inference methodology into the larger framework.

\begin{algorithm}[H]
\caption{Standard Multi-Agent Deep Deterministic Policy Gradient for $N$ agents}
\label{alg:maddpg}
\begin{algorithmic}[1]
\For{episode $= 1$ to $M$}
    \State Initialize a random process $\mathcal{N}$ for action exploration
    \State Receive initial state $x$
    \For{$t = 1$ to max-episode-length}
        \For{each agent $i$}
            \State Select action $a_i = \mu_{\theta_i}(o_i) + \mathcal{N}_t$
        \EndFor
        \State Execute actions $a = (a_1, \dots, a_N)$ and observe reward $r$ and new state $x'$
        \State Store $(x, a, r, x')$ in replay buffer $\mathcal{D}$
        \State $x \gets x'$
        \For{each agent $i$}
            \State Sample minibatch of $S$ samples $(x^j, a^j, r^j, x'^j)$ from $\mathcal{D}$
            \State Set $y^j = r^j_i + \gamma Q_{\phi_i'}(x'^j, a_1', \dots, a_N')$ where $a_k' = \mu_k'(o_k'^j)$
            \State Update critic by minimizing:
            \[
            \mathcal{L}(\phi_i) = \frac{1}{S} \sum_j \left( y^j - Q_{\phi_i}(x^j, a_1^j, \dots, a_N^j) \right)^2
            \]
            \State Update actor using the sampled policy gradient:
            \[
            \nabla_{\theta_i} J \approx \frac{1}{S} \sum_j \nabla_{\theta_i} \mu_i(o_i^j) \nabla_{a_i} Q_{\phi_i}(x^j, a_1^j, \dots, a_i, \dots, a_N^j) \big|_{a_i = \mu_i(o_i^j)}
            \]
        \EndFor
        \State Update target networks for each agent $i$:
        \[
        \theta_i' \gets \tau \theta_i + (1 - \tau) \theta_i', \quad
        \phi_i' \gets \tau \phi_i + (1 - \tau) \phi_i'
        \]
    \EndFor
\EndFor
\end{algorithmic}
\end{algorithm}

\begin{algorithm}[H]
  \caption{Pre-Training Action Inference (Separable Observations)}
  \label{alg:PTAI Training w/ Sep.ble Obs}
  \begin{algorithmic}[1]
    \For{episode $=1$ to $P$}
        \State Receive initial state $\mathbf{x}$ and observation $o$
        \For{$t=1$ to $T$}
            \For{each agent $i$}
                \State $a_i = \text{a\_space.sample}()$
            \EndFor
            \State $\mathbf{o^-} \gets \mathbf{o}$
            \State Execute actions $a=(a_1,\dots,a_N)$, observe reward $r$ and observation $o$
            \For{each agent $i$}
                \State $\Delta o_i = o_i - o^-_i$
                \For{each agent $k$}
                    \If{uniform.random(0,1) $>$ percent-to-train-on}
                        \State \textbf{continue}
                    \EndIf
                    \If{k == i}
                        \State Update Directional-Self-Awareness-Module by minimizing 
                        \[
                        \mathcal{L}(\theta_i) = \text{MSE}(\text{onehot}(a_{i,i}) - \textbf{Self\_Awareness}_{\theta}(o_{i,i},o_{i,g}, o^-_{i,i},o^-_{i,g}, \Delta o_{i,i},\Delta o_{i,g}))
                        \]
                    \Else{}
                        \State Update Directional-Social-Awareness-Module by minimizing 
                        \[
                        \mathcal{L}(\theta_i) = \text{MSE}(\text{onehot}(a_{i,k}) - \textbf{Social\_Awareness}_{\theta}(o_{i,k}, o_{i,i},o_{i,g}, o^-_{i,k}, o^-_{i,i},o^-_{i,g}, \Delta o_{i,k}, \Delta o_{i,i},\Delta o_{i,g}))
                        \]
                    \EndIf
                \EndFor
            \EndFor
        \EndFor
    \EndFor
    \end{algorithmic}
\end{algorithm}

\begin{algorithm}[H]
  \caption{MADDPG with Pre-Trained Action Inference for $N$ agents}
  \label{alg:maddpg w/ PTAI}
  \begin{algorithmic}[1]
    \For{episode $=1$ to $M$}
        \State Initialize a random process $\mathcal{N}$ for action exploration
        \State Receive initial state $\mathbf{x}$ and observation $o$
        \For{each agent $i$}
          \State $a_i \gets \mu_{\theta_i}(\mathbf{o_i}, \vec{0} \in \mathbb{R}^{N\times|\text{a\_space}|}) + \mathcal{N}_t$ 
        \EndFor
        \State $\mathbf{o^-} \gets \mathbf{o}$
        \State Execute actions $a=(a_1,\dots,a_N)$, observe reward $r$, observation $o$, and next state $\mathbf{x}'$
        \State $\mathbf{x} \gets \mathbf{x}'$
      \For{$t = 2$ to max-episode-length}
        \For{each agent $i$}
            \State $\mathbf{\hat{a}_i} = \textbf{AI\_Net}(\mathbf{o_i}, \mathbf{o_i^-})$
            \State $a_i \gets \mu_{\theta_i}(o_i, \mathbf{\hat{a_i}}) + \mathcal{N}_t$ 
        \EndFor
        \State $\mathbf{o^-} \gets \mathbf{o}$
        \State Execute actions $a=(a_1,\dots,a_N)$, observe reward $r$, observation $o$, and next state $\mathbf{x}'$
        \State Store $(\mathbf{x},a,r,\mathbf{x}', o, o^-)$ in replay buffer $\mathcal{D}$
        \State $\mathbf{x}\gets \mathbf{x}'$
        \For{agent $i=1$ to $N$}
          \State Sample a random minibatch of $S$ tuples $(\mathbf{x}^j,a^j,r^j,\mathbf{x}'^j, o^j, o^{-j})$ from $\mathcal{D}$
          \State $\mathbf{\hat{a}_i^{j+1}} = \textbf{AI\_Net}(\mathbf{o_i^{j+1}}, \mathbf{o_i^{j}})$
          \State $y_i^j \gets r_i^j + \gamma\,Q_i^{\mu'}\bigl(\mathbf{x}^{\prime j},a_1',\dots,a_N'\bigr)\bigl|_{a_k'=\mu'_k(o_k^{j+1}, \mathbf{\hat{a}_i^{j+1}})}$
          \State Update critic by minimizing
          \[
            \mathcal{L}(\theta_i)
            = \frac{1}{S}\sum_j \Bigl(y_i^j - Q_i^{\mu}(\mathbf{x}^j,a_1^j,\dots,a_N^j)\Bigr)^{2}
          \]
          \State Update actor using the sampled policy gradient:
          \State $\mathbf{\hat{a}_i^j} = \textbf{AI\_Net}(\mathbf{o_i^j}, \mathbf{o_i^{-j}})$
          \[
            \nabla_{\theta_i}J
            \approx \frac{1}{S}\sum_j
            \nabla_{\theta_i}\mu_i(o_i^j, \mathbf{\hat{a}_i^j})\;
            \nabla_{a_i}Q_i^{\mu}(\mathbf{x}^j,a_1^j,\dots,a_i,\dots,a_N^j)
            \Bigl|_{a_i=\mu_i(o_i^j, \mathbf{\hat{a}_i^j})}
          \]
        \EndFor
        \For{each agent $i$}
          \State Update target networks:
          \[
            \theta_i' \,\gets\, \tau\,\theta_i + (1-\tau)\,\theta_i'
          \]
        \EndFor
      \EndFor
    \EndFor
  \end{algorithmic}
\end{algorithm}

\end{document}